\documentclass[10pt,a4paper]{article}
\usepackage[latin1]{inputenc}
\usepackage{amsmath}
\usepackage{amsfonts}
\usepackage{multicol}
\usepackage{algorithm}
\usepackage{framed}
\usepackage{diagbox}
\usepackage{grffile}
\usepackage[noend]{algpseudocode}
\usepackage{amssymb}
\usepackage{graphicx}
\usepackage{authblk}
\usepackage{booktabs,array}
\usepackage{listings}
\usepackage{stfloats}
\usepackage{url}

\makeatletter
\def\BState{\State\hskip-\ALG@thistlm}
\makeatother

\newcommand*{\QED}[1][$\square$]{%
	\leavevmode\unskip\penalty9999 \hbox{}\nobreak\hfill
	\quad\hbox{#1}%
}

\title{\textbf{Efficient Computation of Hessian Matrices in TensorFlow}}
\author[1,2]{Geir K. Nilsen}
\author[1]{Antonella Z. Munthe-Kaas}
\author[1]{Hans J. Skaug}
\author[1]{Morten Brun}
\affil[1]{Department of Mathematics, University of Bergen}
\affil[2]{geir.kjetil.nilsen@gmail.com}
\date{}                     
\date{}
\allowdisplaybreaks
\renewenvironment{abstract}
{\small
	\begin{center}
		\bfseries \abstractname\vspace{-.5em}\vspace{0pt}
	\end{center}
	\list{}{%
		\setlength{\leftmargin}{0mm}
		\setlength{\rightmargin}{\leftmargin}%
	}%
	\item\relax}
{\endlist}

\begin{document}
\maketitle
\begin{abstract}
This paper deals with the practical aspects of efficiently computing Hessian matrices in the context of deep learning using the Python programming language and the TensorFlow library. We define a general feed-forward neural network model and show how to efficiently compute two quantities: the cost function's exact Hessian matrix, and the cost function's approximate Hessian matrix, known as the Outer Product of Gradients (OPG) matrix. Furthermore, as the number of parameters $P$ in deep learning usually is very large, we show how to reduce the quadratic space complexity by an efficient implementation based on approximate eigendecompositions.
\end{abstract}
\begin{multicols}{2}
\section{Introduction}
The Hessian matrix has a number of important applications in a variety of different fields, such as optimzation, image processing and statistics. Geometrically, the Hessian matrix describes the local curvature of scalar functions $f: \mathbb{R}^P \rightarrow \mathbb{R}$, and is for this reason perhaps mostly known in the field of optimization \cite{nocedal}. Nevertheless, the Hessian matrix also has an important role in statistics, since its inverse is related to the powerful concept of uncertainty quantification \cite{mcfadden}.

In this technical note we mostly focus on the practical aspects of efficiently computing Hessian matrices in the context of deep learning \cite{goodfellowetal} using the Python \cite{python} programming language and the TensorFlow \cite{tensorflow} library. We define a general feed-forward neural network model and show how to efficiently compute two quantities: the cost function's exact Hessian matrix, and the cost function's approximate Hessian matrix, known as the Outer Product of Gradients (OPG) matrix. Furthermore, as the number of parameters $P$ in deep learning usually is very large, we show how to reduce the quadratic space complexity by efficient approximate eigendecompositions. Although we here use a feed-fordward neural network architecture to introduce terminology, the theory and implementation presented is still directly applicable on more general neural network architectures using convolutional layers, pooling and regularization.

The paper is organized as follows: In Section \ref{section:definitions} we give definitions which will be used throughout the paper. In Section \ref{section:estimatingthehessian} we present the problem statement, and  discuss three complications which need to dealt with in order to achieve a successful TensorFlow implementation: 1) \texttt{tf.hessians()} is fundamentally inadequate since it only calculates a subset of all the partial derivatives (Section \ref{section:tfhessians}), 2) computing Hessian matrices essentially requires per-example gradients of the cost function with respect to model parameters, and unfortunately, the differentiation functionality provided by TensorFlow does not support computing gradients with respect to individual examples efficiently \cite{goodfellow} (Section \ref{section:per-example}), and 3) when differentiating a function with respect to several variables represented by a list of tensors, the result is also a list of tensors (Section \ref{section:gradrep}). In Section \ref{section:hessalg} we show how to overcome the aforementioned complications and introduce our Python module \texttt{pyhessian} \cite{pyhessianmodule} which is released as open source licensed under GNU GPL on GitHub. In Section \ref{section:summary} we summarize the paper and give some concluding remarks.
\end{multicols}
\section{Deep Neural Networks}\label{section:definitions}
\begin{figure*}[h]
	\centering
	\includegraphics[scale=0.36]{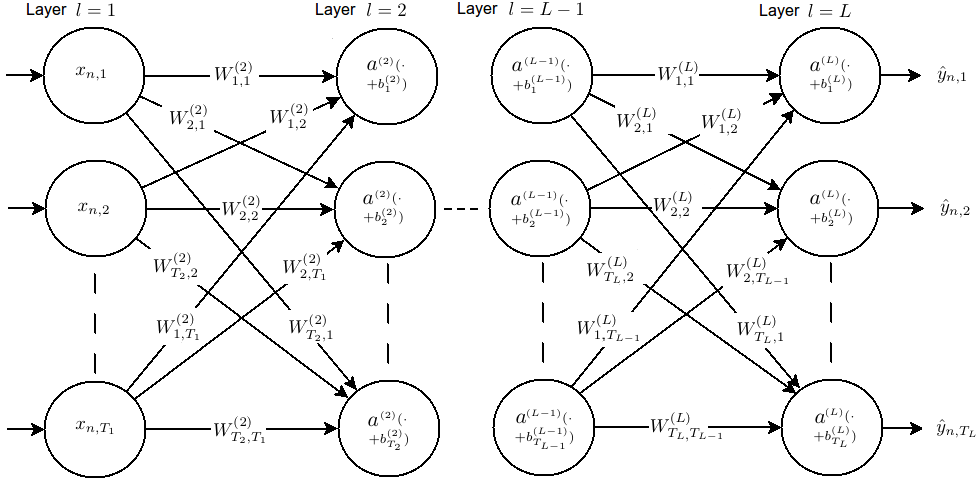}
\caption{A Feed-Forward Neural Network with Dense Layers}
	\label{fig:ffnn}
\end{figure*}

A feed-forward neural network is shown in Figure (\ref{fig:ffnn}). There are $L$ layers $l=1,2,...,L$ with $T_l$ neurons in each layer. The input layer $l=1$, is represented by the input vector $x_n = \begin{bmatrix}x_{n,1} & x_{n,2} & \hdots & x_{n,T_1}\end{bmatrix}^T$ where $n=1,2,...,N$ is the input index. Furthermore, there are $L-2$ dense hidden layers, $l=2,3,...,L-1$, and a dense output layer $l=L$, all represented by weight matrices $W^{(l-1)} \in \mathbb{R}^{T_{l} \times T_{l-1}}$, bias vectors $b^{(l)} \in {\mathbb{R}^{T_l}}$ and vectorized activation functions $\sigma^{(l)}$.
\begin{multicols}{2}
\noindent Let the cost function $C$ coincide with TensorFlow's built-in softmax cross-entropy function\footnote{TensorFlow API r1.13: tf.losses.softmax\_cross\_entropy()},
\begin{align}
C &= \frac{1}{N}\sum_{n=1}^{N}C_n(y_{n}, \hat{y}_n)\label{eq:cost}\\ &= \frac{1}{N}\sum_{n=1}^N \left( -\sum_{m=1}^{T_L} y_{n,m} \text{log } \hat{y}_{n,m} \right).
\end{align}
It is defined as the average of $N$ per-example cross-entropy cost functions~$C_n(y_{n}, \hat{y}_n)$, where $y_{n}$ represents the one-hot target vector for the $n$th example, and where $\hat{y}_{n}$ represents the corresponding prediction vector. The prediction vector is obtained by evaluating the model function (\ref{eq:modelfunc}) 
\begin{figure*}[t]
	\begin{equation}\label{eq:modelfunc}
		\hat{y}_{n} = f(x_n, \omega) = \sigma^{(L)}( W^{(L-1)}\sigma^{(L-1)}( \cdots \sigma^{(2)}(W^{(1)}x_n + b^{(2)}) + \cdots ) + b^{(L)})
	\end{equation}
\end{figure*}
using the input vector $x_n$ and a flat vector of model parameters $\omega \in \mathbb{R}^P$ defined by
\begin{align}
	\omega &= \begin{bmatrix}\omega_1 & \omega_2 & \hdots & \omega_P\end{bmatrix}^T\\ &= \underset{l=2,3,\hdots,L}{\text{flatten}}(W^{(l-1)},b^{(l)}).\label{eq:flatten}
\end{align}
The function $\text{flatten}(\cdot)$ denotes a row-wise flattening operation to transform the collection of model parameters represented by the weight matrices $W^{(l-1)}$ and bias vectors $b^{(l)}, l=2,3,...,L$ into a flat column vector of dimension $P = T_1 T_2 + T_2 + \hdots + T_{L-1}T_L + T_L$. Further, the activation function in the output layer is the vectorized softmax function
\begin{align}
\sigma^{(L)}(z) &= \text{softmax}(z)\\ &= \frac{\text{exp}{(z)}}{\sum_{m=1}^{T_L}\text{exp}{(z_{m})}},
\end{align}
where $z \in \mathbb{R}^{T_L}$, and where $\text{exp}(\cdot)$ denotes the vectorized exponential function. Finally, training of the neural network can be defined as finding an `optimal' parameter vector $\hat{\omega}$ by minimizing the cost function (\ref{eq:cost}),
\begin{equation}
	\hat{\omega} = \underset{\omega \in \mathbb{R}^{P}}{\text{arg min}~C(\omega)}.\label{eq:omegahat}
\end{equation}
\section{Computing Hessian Matrices in TensorFlow}\label{section:estimatingthehessian}
Given the cost function $C$ defined in Section \ref{section:definitions}, the Hessian matrix $H \in \mathbb{R}^{P\times P}$ is defined\footnote{The notation used means that $H_{i,j}=\frac{\partial^2 C}{\partial \omega_i \partial \omega_j}\bigg\rvert_{\overset{\omega_i=\hat{\omega}_i}{\omega_i=\hat{\omega}_i}}$}
\begin{align}
H &=  \frac{\partial^2 C}{\partial \omega \partial \omega^T}\bigg\rvert_{\omega=\hat{\omega}}\\&= \frac{1}{N}\sum_{n=1}^N
	 \frac{\partial^2 C_n}{\partial \omega \partial \omega^T}\bigg\rvert_{\omega=\hat{\omega}}\label{eq:hesseq1}.
\end{align}
The approximation to the Hessian matrix, known as the Outer Product of Gradients (OPG) matrix $G \in \mathbb{R}^{P \times P}$, is defined
\begin{align}
	G &= \frac{1}{N}\sum_{n=1}^N \frac{\partial C_n}{\partial \omega} \frac{\partial C_n}{\partial \omega}^T\bigg\rvert_{\omega=\hat{\omega}}\label{eq:hesseq2}\\&\neq\frac{\partial C}{\partial \omega} \frac{\partial C}{\partial \omega}^T\bigg\rvert_{\omega=\hat{\omega}}\label{eq:hesseq4}.
\end{align}
Letting $J = \begin{bmatrix}\frac{\partial C_1}{\partial \omega} & \frac{\partial C_2}{\partial \omega} & \hdots & \frac{\partial C_N}{\partial \omega} \end{bmatrix}$, yields
\begin{align}
	G &= \frac{1}{N}J^T J\bigg\rvert_{\omega=\hat{\omega}}\label{eq:hesseq3}.
\end{align}
We notice that $H$ in Equation (\ref{eq:hesseq1}) is formed by summing over $N$ per-example Hessian matrices, and that $G$ in Equation (\ref{eq:hesseq2}) is formed by summing over $N$ per-example OPG matrices. We also note that $H$ can be obtained by differentating the cost function directly, whereas this property does not hold for $G$ as seen by (\ref{eq:hesseq4}). Finally, we note that $G$ can be written as a per-example cost Jacobian matrix product (\ref{eq:hesseq3}).

In order to proceed, we now need to consider three complications regarding gradients and Hessians in TensorFlow: the limitations of TensorFlow's built-in \texttt{tf.hessians()} function is discussed in Section \ref{section:tfhessians}, per-example gradients will be discussed in Section \ref{section:per-example}, and gradient representation will be discussed in Section \ref{section:gradrep}.
\subsection{Per-Example Gradients}\label{section:per-example}
A per-example gradient of the cost function with respect to model parameters means to differentiate $C_n$ in (\ref{eq:hesseq1}) and (\ref{eq:hesseq2}) with respect to model parameters for a single example $n$. However, when TensorFlow compute gradients (e.g. \texttt{tf.gradients()}) it performs back propagation, which never actually computes the per-example gradients, but instead directly obtains the sum of per-example gradients. To see what this means, consider the following dummy multiple linear regression model (for simplicity with no bias term):
\end{multicols}
\begin{lstlisting}[frame=tb]
In [1]: import tensorflow as tf
In [2]: import numpy as np
In [3]: W = tf.Variable([3., 4., 5., 2.])
In [4]: X = tf.placeholder('float32', shape=(None, 4))
In [5]: yhat = tf.tensordot(X, W, axes = 1)
In [6]: init = tf.global_variables_initializer()
In [7]: sess = tf.InteractiveSession()
In [8]: sess.run(init)
\end{lstlisting}
\noindent We have model parameters represented by the variable tensor \texttt{W} (\texttt{In [3]}), and we use the placeholder tensor \texttt{X} (\texttt{In [4]}) as the model input. For simplicity, we do not define a cost function here, but instead conduct several differentiation experiments directly on the scalar model function \texttt{yhat} (\texttt{In [5]}) with $N=2$:
\begin{lstlisting}[frame=tb]
 In [9]: sess.run(yhat, feed_dict={x:np.array([[1.,2.,3.,4.], 
                                               [2.,3.,4.,5.]])})
Out [1]: array([34., 48.], dtype=float32)
\end{lstlisting}
\noindent We get back two values (\texttt{Out [1]}) corresponding to the two inner products as expected. We now take the gradient of the model function with respect to the model parameters for a single example:
\begin{lstlisting}[frame=tb]
In [10]: sess.run(tf.gradients(yhat, W), 
                  feed_dict={X:np.array([[1.,2.,3.,4.]])})
Out [2]: [array([1., 2., 3., 4.], dtype=float32)]
\end{lstlisting}
\noindent We get back the per-example gradient as expected (\texttt{Out [2]}). We do the same for the second example:
\begin{lstlisting}[frame=tb]
In [11]: sess.run(tf.gradients(yhat, W), 
                  feed_dict={X:np.array([[2.,3.,4.,5.]])})
Out [3]: [array([2., 3., 4., 5.], dtype=float32)]
\end{lstlisting}
\noindent But when we try to feed two examples:
\begin{lstlisting}[frame=tb]
In [12]: sess.run(tf.gradients(yhat, W), 
                  feed_dict={X:np.array([[1.,2.,3.,4.], 
                                         [2.,3.,4.,5.]])})
Out [4]: [array([3., 5., 7., 9.], dtype=float32)]
\end{lstlisting}
\noindent we notice that we do not get back two per-example gradients, but rather the sum of the two per-example gradients (\texttt{Out [4]}). The important observation is here that in order to obtain per-example gradients we seemingly need to run \texttt{tf.gradients()} once per example, which in turn is well known to be very inefficient when $N$ grows large. We will get back to this and discuss solutions in Sections (\ref{section:estimatingH}) and \ref{section:estimatingG}).
\begin{multicols}{2}
\subsection{Gradient Representation}\label{section:gradrep}
In practice, the $P$ model parameters are represented by a list of tensors (e.g. \texttt{[tf.Variable(),...]}) corresponding to the different layers of the model architecture. On the other hand, the Hessian matrix is only one \texttt{(P, P)}-shaped tensor (matrix) formed by every single variable element contained in the list of variable tensors.

When differentiating a function represented by a computational graph with respect to some variable(s) in that graph, the variable tensors we pass to the differentiation function (\texttt{tf.gradients()}) must be kept in their original form as upon defining the graph. One can still pass on the whole collection of variables as a list to get hold of the full gradient, but the result will not be a flat gradient vector -- it will rather be a list of sub-gradients represented by multiple tensors. This means that in order to end up with the \texttt{(P, P)}-shaped Hessian matrix we want, we need to keep all the variables in a list during differentiation, and only afterwards reshape the result into the desired flat form.
\end{multicols}
\subsubsection{Flattening of Gradients}\label{section:flattening}
To illustrate the concept of lists of sub-gradients vs. flat gradients, consider a dummy multinomial logistic regression model:
\begin{lstlisting}[frame=tb]
In [13]: import tensorflow as tf
In [14]: T1 = 64
In [15]: T2 = 32
In [16]: P = T1*T2 + T2 # Total number of model parameters
In [17]: W = tf.Variable(tf.ones((T1, T2)), 'float32')
In [18]: b = tf.Variable(tf.ones((T2,)), 'float32')
In [19]: params = [W, b]
In [20]: params
Out [5]: [<tf.Variable 'Variable...' shape=(64, 32) ...>,
          <tf.Variable 'Variable...' shape=(32,) ...]
In [21]: X = tf.placeholder(dtype='float32', shape=(None, T1))
In [22]: y = tf.placeholder(dtype='float32', shape=(None, T2))
In [23]: def model_fun(X, params):
             return tf.add(tf.matmul(X, params[0]), params[1])
In [24]: yhat_logits = model_fun(X, params)
In [25]: yhat = tf.nn.softmax(yhat_logits)
In [26]: def cost_fun(y, yhat_logits, params):
             return tf.losses.softmax_cross_entropy(y, 
                                                    yhat_logits)
In [27]: cost = cost_fun(y, yhat_logits, params)
\end{lstlisting}
\noindent We thus have model parameters \texttt{W} (\texttt{In [17]}) and \texttt{b} (In [18]) with shapes \texttt{(T1, T2)} and \texttt{(T2,)}, respectively. We can differentiate the cost function represented by the tensor \texttt{cost} (\texttt{In [27]}) with respect to the individual variables, or the full list \texttt{params} (\texttt{In [19]}):
\begin{lstlisting}[frame=tb]
In [28]: tf.gradients(cost, W)
Out [6]: [<tf.Tensor 'gradients...' shape=(64, 32) ...>]
In [29]: tf.gradients(cost, b)
Out [7]: [<tf.Tensor 'gradients...' shape=(32,) ...>]
In [30]: tf.gradients(cost, params)
Out [8]: [<tf.Tensor 'gradients...' shape=(64, 32) ...>,
          <tf.Tensor 'gradients...' shape=(32,) ...>]
\end{lstlisting}
\noindent But if we try to reshape our parameters into a flat vector and then differentiate:
\begin{lstlisting}[frame=tb]
 In [31]: params_flat = tf.concat([tf.reshape(W, [-1]), b], 
                                  axis=0)
 In [32]: params_flat
 Out [9]: <tf.Tensor 'concat...' shape=(2080,) ...>
 In [33]: tf.gradients(cost, params_flat)
Out [10]: [None]
\end{lstlisting}
\noindent We get \texttt{[None]} (\texttt{Out [10]}) because the new tensor \texttt{params\_flat} (\texttt{In [31]}) is not part of the \texttt{cost} function graph (\texttt{In [27]}). We solve the issue by first differentiating with respect to the full list, and then flattening the resulting tensor:
\begin{lstlisting}[frame=tb]
 In [34]: grads = tf.gradients(cost, params)
 In [35]: grads
Out [12]: [<tf.Tensor 'gradients_...' shape=(64, 32) ...>,
           <tf.Tensor 'gradients_...' shape=(32,) ...>]
 In [36]: grads_flat = tf.concat([tf.reshape(grads[0],[-1]), 
                                  grads[1]], 
                                 axis=0)
 In [37]: grads_flat
Out [13]: <tf.Tensor 'concat...' shape=(2080,) dtype=float32>
\end{lstlisting}
\subsection{The built-in TensorFlow function \texttt{tf.hessians()}}\label{section:tfhessians}
The fundamental question is, why can we not simply use the built-in TensorFlow function \texttt{tf.hessians()}? To see why, consider the following:
\begin{lstlisting}[frame=tb]
 In [38]: tf.hessians(cost, params)
Out [14]: [<tf.Tensor 'Reshape_...' shape=(64, 32, 64, 32) ...>,
           <tf.Tensor 'Reshape_...' shape=(32, 32) ...>]
\end{lstlisting}
\begin{multicols}{2}
\noindent We observe that we get back two tensors (\texttt{Out [14]}). Let us name the two $H_U$ and $H_L$, respectively. Their respective shapes are \texttt{(T1, T2, T1, T2)} and \texttt{(T2, T2)}. Firstly, if we reshape $H_U$ into a \texttt{(T1*T2, T1*T2)}-shaped tensor, it will correspond to the full Hessian's upper block diagonal matrix $\in \mathbb{R}^{T_1 T_2 \times T_1T_2}$. Secondly, the tensor $H_L$ corresponds to the full Hessian's lower block diagonal matrix $\in \mathbb{R}^{T_2 \times T_2}$. In other words, we get no information about the full Hessian's two off-diagonal block matrices $\in \mathbb{R}^{T_1 T_2 \times T_2}$ and $\mathbb{R}^{T_2 \times T_1T_2}$. 
Equation (\ref{eq:tf.hess.trouble}) illustrates the concept.
\setlength{\abovedisplayskip}{10pt}
\begin{equation}
H = \begin{bmatrix}
H_U \in \mathbb{R}^{T_1 T_2 \times T_1 T_2} & ? \in \mathbb{R}^{T_1T_2 \times T_2}\\
? \in \mathbb{R}^{T_2 \times T_1T_2} & H_L \in \mathbb{R}^{T_2 \times T_2} \\
\end{bmatrix}
\label{eq:tf.hess.trouble}
\end{equation}
The two missing off-diagonal block matrices\footnote{The two matrices are equal up to transposition, since $H$ is symmetric} represented by question marks in Equation (\ref{eq:tf.hess.trouble}) correspond to the partial derivatives involving variable entities from different tensors in the parameter list \texttt{params} (\texttt{In [19]}). The same principle applies for all \texttt{params} with $\texttt{len(params)} > 1$.

\section{Approximate Hessian Eigendecompositions}\label{sec:eigdecomphessian}
In deep learning, the number of parameters $P$ is usually so large that the full Hessian matrix will be prohibitively expensive to compute and store. In this section we present methodology addressing the issue in terms of approximate eigendecompositions based on $K$ eigenpairs. Thus leading to a space complexity of $O(KP)$ rather than $O(P^2)$. As the time complexity is somewhat more involved, we leave this discussion for Sections \ref{sec:computingeigdecomphessian} and \ref{sec:computingeigdecompopg}.
\subsection{Low-rank Approximation}
A low-rank approximation of the Hessian matrix can be obtained by a eigendecomposition utilizing only $K$ eigenpairs corresponding to the $K$ largest eigenvalues of $H$ (or $G$),
\begin{equation}
	\widetilde{H} = Q\Lambda Q^T\label{eq:lowrankapprox} \in \mathbb{R}^{P \times P},
\end{equation}
where $Q \in \mathbb{R}^{P \times K}$ is the matrix whose $k$th column is the eigenvector $q_k$ of $H$ (or $G$), and $\Lambda \in \mathbb{R}^{K \times K}$ is the diagonal matrix whose elements are the corresponding eigenvalues, $\Lambda_{kk} = \lambda_k$. We assume that the eigenvalues are algebraically sorted so that $\lambda_1 \ge \lambda_2 \ge \hdots \ge \lambda_K$.
\subsection{Full-rank Approximation}
A full-rank approximation of the Hessian matrix can be obtained by an extrapolation of its smallest eigenvalues. Assuming that $\lambda_{K+1} = \lambda_{K+2} = \hdots = \lambda_{P} = \widetilde{\lambda} > 0$, a full-rank approximation is given by
\begin{equation}
	\widetilde{\widetilde{H}} = \widetilde{H} + \widetilde{\lambda}(I - Q Q^T) \in \mathbb{R}^{P \times P},
\end{equation}
where $\widetilde{H}$ is the low-rank approximation \eqref{eq:lowrankapprox} and where we have used that $Q$ is an orthonormal basis. Details can be found in the Appendix \ref{sec:full-rankproof}. One particular choice for $\widetilde{\lambda}$ is to set it equal to the smallest eigenvalue in the low-rank approximation, e.g. $\widetilde{\lambda}=\lambda_K$.
\section{Implementation}\label{section:hessalg}
We will now address how to overcome the basic complications discussed in Sections \ref{section:tfhessians}, \ref{section:per-example} and \ref{section:gradrep}. The current section is divided into four parts: we first discuss how to compute the matrix $H$ in Equation (\ref{eq:hesseq1}), and afterwards move on to the matrix $G$ in Equation (\ref{eq:hesseq2}). Finally, in Sections \ref{sec:computingeigdecomphessian} and \ref{sec:computingeigdecompopg} we address how to compute the aforementioned approximate eigendecompositions of both $H$ and $G$.

\subsection{Computing $H$}\label{section:estimatingH}
We compute the matrix $H$ based on Hessian vector products \cite{pearlmutter}. A practial implementation of Equation (\ref{eq:hesseq1}) is essentially to form $P$ Hessian vector products using the full set of basis vectors in $\mathbb{R}^P$. As a bonus, the resulting implementation can easily be paralellized because the columns of the Hessian matrix can be computed independently.

In the following we describe the essential parts of this paper's accompanying Python module \texttt{pyhessian} \cite{pyhessianmodule}.
The Hessian vector product function \texttt{get\_Hv\_op(v)} can be described as follows:
\begin{enumerate}
	\item Differentiates the cost function with respect to the model parameters contained in the list \texttt{params} and flattens the result
	\item Performs elementwise multiplication of the flattened gradient and the vector \texttt{v}; \texttt{tf.stop\_gradient()} ensures that \texttt{v} is treated as a constant during differentiation. This is important if the vector \texttt{v} is a function of the model parameters $\omega$.
	\item Differentiates the resulting elementwise vector product with respect to the model parameters (to get second order derivatives) and flattens the result. As this step can appear subtle, see the Appendix \ref{sec:hessvecproof} for a rigorous derivation.
\end{enumerate}
Note that the function \texttt{get\_Hv\_op(v)} uses the function \texttt{flatten()} which is based on the insights from Section \ref{section:flattening} and the mathematical operation defined in Equation (\ref{eq:flatten}).
Furthermore, we have defined a parallellized function \texttt{get\_H\_op()} to create the full Hessian matrix operation based on forming \texttt{P} Hessian vector products using \texttt{get\_Hv\_op(v)} for all $v$'s in $\mathbb{R}^P$. The function \texttt{get\_H\_op()} sets up a parallellized operation using \texttt{tf.map\_fn()} to get hold of all the \texttt{P} columns of the full Hessian matrix as defined in Equation (\ref{eq:hesseq1}). It works by applying \texttt{Hv\_op} on all basis vectors in $\mathbb{R}^P$ represented by \texttt{tf.eye(self.P, self.P)}, where \texttt{P} is the total number of parameters in the model.

The important remark is now to realize that, by definition, the matrix $H$ in Equation (\ref{eq:hesseq1}) is the sum of per-example Hessian matrices. It means that we can directly leverage from the fact that \texttt{tf.gradients()} returns the sum of per-example gradients discussed in Section \ref{section:per-example}. 
In other words, when we run the resulting \texttt{H\_op} in a graph session, we get per-example Hessians (below \texttt{In [43]}) if we feed single examples, and the average of per-example Hessians if we feed more than one example. Thus, we can get a mini-batch (below using size \texttt{batch\_size\_H}) Hessian matrix if we feed a mini-batch (below \texttt{In [45]}), or we can obtain the full Hessian matrix directly by feeding the complete training set. However, to avoid excessive memory consumption for large $N$, we can sum over mini-batch Hessians and divide by the number of mini-batches (\texttt{In [46] - In [56]}):
\end{multicols}
\begin{lstlisting}[basicstyle=\scriptsize, frame=tb, caption={Computing $H$},label={lst:label1}]
In [39]: from pyhessian import HessianEstimator
In [40]: hest = HessianEstimator(...)
In [41]: H_op = hest.get_H_op()
In [42]: # Per-example
In [43]: H = sess.run(H_op, feed_dict={X:[X_train[0]],
                                       y:[y_train[0]]}) 
In [44]: # Mini-batch
In [45]: H = sess.run(H_op, feed_dict={X:X_train[:batch_size_H],
                                       y:y_train[:batch_size_H]})
In [46]: # Full
In [47]: B = int(N/batch_size_H)
In [48]: H = np.zeros((hest.P, hest.P), dtype='float32')
In [49]: for b in range(B): 
In [50]:     H = H + sess.run(H_op, 
In [51]:                      feed_dict={ \
In [52]:                      X: X_train[b*batch_size_H: \
In [53]:                                 (b+1)*batch_size_H], 
In [54]:                      y: y_train[b*batch_size_H: \
In [55]:                                 (b+1)*batch_size_H]})
In [56]: H = H/B
\end{lstlisting}
\begin{multicols}{2}
\subsection{Computing $G$}\label{section:estimatingG}
Due to the inequality sign in Equation (\ref{eq:hesseq4}), the computation of $G$ (unlike $H$) cannot exploit the implicit sum of gradients as discussed in Section \ref{section:per-example}. Instead, we will pursuit another efficient technique based on parallized per-example gradients. Although the technique we present here has been reformulated and adapted to our needs, the original implementation idea is to our knowledge originating from the author of \cite{goodfellow}. 
The OPG matrix operation function \texttt{get\_G\_op()} can be described as follows:
\begin{enumerate}
   	\item Creates \texttt{batch\_size\_G} copies of the model parameters
   	\item Splits the model input variable \texttt{X}, and the model output variable \texttt{y} into respective lists of \texttt{batch\_size\_G} elements
    \item Creates a list of \texttt{batch\_size\_G} elements holding model output tensors resulting from evaluating the model function using respective inputs and parameter copies
    \item Creates a list of \texttt{batch\_size\_G} elements holding cost output tensors resulting from evaluating the cost using respective labels, model outputs and parameter copies
    \item Stacks up a flat per-example gradient tensor by paralell differentiation of per-example costs with respect to the corresponding model parameter copy
    \item Forms the OPG matrix operation by matrix multiplication of per-example cost Jacobians as in Equation (\ref{eq:hesseq3})
\end{enumerate}
Note that the function \texttt{get\_G\_op()} utilizes the function \texttt{flatten()} which is based on the insights from Section \ref{section:flattening} and the mathematical operation defined in Equation (\ref{eq:flatten}). Also note that the function \texttt{get\_G\_op()} requires itself to maintain redundant model parameter copies which size scale with \texttt{batch\_size\_G}. To avoid excessive memory consumption, we can sum over mini-batch OPGs and divide by the number of mini-batches (\texttt{In [64]} - \texttt{In [68]}):
\end{multicols}
\begin{lstlisting}[basicstyle=\scriptsize, frame=tb, caption={Computing $G$},label={lst:label2}]
In [57]: hest = HessianEstimator(..., batch_size_G)
In [58]: G_op = hest.get_G_op()
In [59]: # Per-example
In [60]: sess.run(G_op, feed_dict={X:[X_train[0]], 
                                   y:[y_train[0]]})
In [61]: # Mini-batch
In [62]: sess.run(G_op, feed_dict={X:X_train[:batch_size_G],
                                   y:y_train[:batch_size_G]}) 
In [63]: # Full
In [64]: B = int(N/batch_size_G)
In [65]: G = np.zeros((hest.P, hest.P), dtype='float32')
In [66]: for b in range(B): 
In [67]:     G = G + sess.run(G_op, 
                              feed_dict={ \
                              X: X_train[b*batch_size_G:\
                                         (b+1)*batch_size_G], 
                              y: y_train[b*batch_size_G:\
                                         (b+1)*batch_size_G]})
In [68]: G = G/B
\end{lstlisting}
\begin{multicols}{2}
\subsection{Computing Eigenpairs of $H$}\label{sec:computingeigdecomphessian}
The Lanczos iteration \cite{trefethen} can be applied to find $K < P$ eigenvalues (and corresponding eigenvectors) in $O(SNP)$ time and $O(KP)$ space when Pearlmutter's technique \cite{pearlmutter} is applied inside the iteration. Pearlmutter's technique can simply be described as a procedure based on two-pass back-propagations of complexity $O(NP)$ time and $O(P)$ space to obtain exact Hessian vector products without requiring to keep the full Hessian matrix in memory. The number $S$ denotes the number of Lanczos iterations to reach convergence. Typically the convergence of the Lanczos algorithm will be fast enough so that $S$ is orders of magnitude less than $P$.

Essentially, we select the number of eigenapirs $K$ and use \texttt{LinearOperator} from the \texttt{scipy} distribution in combination with the Lanczos implementation \texttt{eigsh}, and setup the former to compute Hessian vector products using \texttt{get\_Hv\_op()} from \texttt{pyhessian} (\texttt{In [69]} - \texttt{In [74]}). The \texttt{LinearOperator} (\texttt{In [83]}) is initialized with a callback function \texttt{Hv()} (\texttt{In [75] - \texttt{In [82]}}) where the actual graph session is executed. The \texttt{eigsh} argument \texttt{which='LA'} (\texttt{In [84]}) ensures that the eigenpairs returned corresponds to the algebraically largest eigenvalues of $H$, and the lines \texttt{In [85]} - \texttt{In [87]} sorts the eigenpairs in descending eigenvalue order.
\end{multicols}
\begin{lstlisting}[basicstyle=\scriptsize, frame=tb, caption={Computing the Eigendecomposition of $H$},label={lst:label3}]
In [69]: from scipy.sparse.linalg import LinearOperator
In [70]: from scipy.sparse.linalg import eigsh
In [71]: K = 10
In [72]: hest = HessianEstimator(...)
In [73]: _v = tf.placeholder(shape=(hest.P,), dtype='float32')
In [74]: Hv_op = hest.get_Hv_op(_v)
In [75]: def Hv(v):
In [76]:     B = int(N/batch_size_H)
In [77]:     Hv = np.zeros((hest.P))
In [78]:     Bs = batch_size_H
In [79]:     for b in range(B):
In [80]:         Hv = Hv + sess.run(Hv_op, 
                                    feed_dict={X:X_train[b*Bs:\
                                                         (b+1)*Bs], 
     	                                       y:y_train[b*Bs:\
                                                         (b+1)*Bs], 
     	                                       _v:np.squeeze(v)})
In [81]:     Hv = Hv / B
In [82]:     return Hv
In [83]: H = LinearOperator((hest.P, hest.P), matvec=Hv, 
                            dtype='float32')
In [84]: L, Q = eigsh(H, k=K, which='LA')
In [85]: sinds = np.flip(np.argsort(L))
In [86]: L = L[sinds]
In [87]: Q = Q[:,sinds]
\end{lstlisting}
\begin{multicols}{2}
\subsection{Computing Eigenpairs of $G$}\label{sec:computingeigdecompopg}
For the OPG approximation \eqref{eq:hesseq4}, a slightly different approach can be applied. Since the OPG matrix can be written as a Jacobian matrix product \eqref{eq:hesseq3}, we get by the singular value decomposition that its eigenvectors will be the right singular vectors of the Jacobian, and its eigenvalues the squared singular values
\begin{align}
	NG = J^T J &= (U\Sigma V^T)^T U\Sigma V^T\nonumber\\
	          &= V \Sigma U^T U \Sigma V^T\nonumber\\
	          &= V \Sigma^2 V^T\label{eq:J}
\end{align}
However, even the $N \times P$-dimensional Jacobian matrix $J$ is prohibitively expensive to store. Luckily,
mini-batches of $J$ can easily be obtained, and so an incremental singular value decomposition \cite{levy, cardot} can be applied to each mini-batch. The computational cost is thus $O(KNP)$ time and $O(KP)$ space. We select the number of eigenapirs $K$ and use \texttt{IncrementalPCA} from the \texttt{sklearn} distribution (\texttt{In [88]} - \texttt{In [94]}). We then make use of $J$ in \eqref{eq:J} which is available via the function \texttt{get\_J\_op()}. The \texttt{get\_J\_op()} implementation is similar to \texttt{get\_G\_op()} except from that it excludes the final matrix product $J^T J$ and just returns $J$ (\texttt{In 95}). Essentially, the rest of the details are tied to filling up the buffer \texttt{J} in a mini-batch fashion and also ensuring that the number of examples per mini-batch is large enough to support the selected $K$ (\texttt{In [96]} - \texttt{In [103]}). Finally, the eigenpairs are computed based on \eqref{eq:J} (\texttt{In [104]}).
\end{multicols}
\begin{lstlisting}[basicstyle=\scriptsize, frame=tb, showstringspaces=false, caption={Computing the Eigendecomposition of $G$},label={lst:label4}]
In  [88]: from sklearn.decomposition import IncrementalPCA
In  [89]: K = 10
In  [90]: Bs = batch_size_G
In  [91]: hest = HessianEstimator(...)
In  [92]: _N = int(np.ceil(K / Bs))
In  [93]: assert N % _N != 0, 'N must be divisible by \
                               K/batch_size_G!'
In  [94]: ipca = IncrementalPCA(n_components=K, batch_size=Bs*N, 
                                copy=False)
In  [95]: J_op = hest.get_J_op()
In  [96]: J = np.zeros((Bs*_N, hest.P), dtype='float32')
In  [97]: B = int(N/Bs)
In  [98]: for b in range(B):
In  [99]:     s1 = Bs*(b%_N)
In [100]:     s2 = Bs*(b%_N+1)
In [101]:     J[s1:s2] = sess.run(J_op, 
                                 feed_dict={X: X_train[b*Bs:\
                                                       (b+1)*Bs], 
                                            y: y_train[b*Bs:\
                                                       (b+1)*Bs]})
In [102]:     if (b+1) % _N == 0:
In [103]:         ipca.partial_fit(J)
In [104]: L, Q = np.float32(ipca.singular_values_**2 / N),\
                 np.float32(ipca.components_.T)
\end{lstlisting}
\begin{multicols}{2}
\subsection{Low-Rank Approximations}
Given the implementations of the eigendecompositions of $H$ and $G$ in Sections \ref{sec:computingeigdecomphessian} and \ref{sec:computingeigdecompopg}, low-rank approximations can be computed by
\begin{lstlisting}[frame=tb, caption={Computing the Low-Rank Approximation},label={lst:label5}]
Q@np.diag(L)@Q.T
\end{lstlisting}
However, the primary motivation of this approximation is to avoid storing the full Hessian in memory. For example, if the intent is to evaluate $y = x^THx$ for $x\in\mathbb{R}^P$ then we can use
\begin{lstlisting}[frame=tb, caption={Implicit Application of the Low-Rank Approximation},label={lst:label6}]
y = (x.T@Q)@np.diag(L)@(Q.T@x)
\end{lstlisting}
where we have intentionally introduced superfluous parenthesis to illustrate that this expression avoids to form a full $P\times P$ matrix as an intermediate step.
\subsection{Full-Rank Approximations}
Given the implementations of $H$ and $G$ in Sections \ref{sec:computingeigdecomphessian} and \ref{sec:computingeigdecompopg}, full-rank approximations (using $\widetilde{\lambda} = \lambda_K$) can be computed by
\begin{lstlisting}[frame=tb, caption={Computing the Full-Rank Approximation}, label={lst:label7}]
Q@np.diag(L)@Q.T \
+ L[-1]*(np.eye(hest.P) \
         - Q@Q.T)
\end{lstlisting}
Analogously to the low-rank example, if we wish to evaluate $y = x^THx$ using the full-rank approximation with no intermediate formation of the full Hessian (nor $I$), we can use
\begin{lstlisting}[frame=tb, caption={Implicit Application of the Full-Rank Approximation}, label={lst:label8}]
y = (x.T@Q)@np.diag(L)@(Q.T@x)\
    + L[-1]*x.T@x \
    - L[-1]*(x.T@Q)@(Q.T@x)
\end{lstlisting}
\section{Summary and Concluding Remarks}\label{section:summary}
We have presented a practical and efficient TensorFlow implementation for computing Hessian matrices in a deep learning context. The naive methods have a complexity of $O(NP^2)$ time and $O(P^2)$ space where $N$ is the number of examples in the training set and $P$ is the number of parameters in the model. Furthermore, we have introduced means for efficient computation of approximate Hessian eigendecompositions based on $K$ eigenpairs, and shown how these can be applied as both low-rank and full-rank operators. The complexity of the approximate eigendecompositon of the Hessian is $O(SNP)$ and $O(KP)$ space where $S$ represents the number of required Lanczos steps, whereas for the OPG approximation $O(KPN)$ time and $O(KP)$ space. The novelty of the naive methodology presented prominently lies in the implementation technique rather than in the asymptotic bound analysis point of view. As noted by \cite{goodfellow}, a naive method running back propagation $N$ times with a mini-batch of size $1$ is very inefficient because TensorFlow's back propagation implementation will not be able to exploit the parallelism of mini-batch operations by efficient matrix operation implementations. An usage example of the \texttt{pyhessian} module \cite{pyhessianmodule} applied on a feed-forward neural network TensorFlow model can be found in the included file \texttt{pyhessian\_example.py}.
\end{multicols}
\section{Appendix}
\subsection{Derivation of the Hessian Vector Product Implementation}\label{sec:hessvecproof}
Let $y$ be the product between the Hessian matrix and an arbitrary vector $v$,
\begin{equation}
	y = v^TH(\omega)|_{\omega=\hat{\omega}} \in\mathbb{R}^{P},
\end{equation}
\begin{equation}
	H(\omega)|_{\omega=\hat{\omega}} = \begin{bmatrix} 
		\frac{\partial^2 C(\omega)}{\partial^2 \omega_1} & \frac{\partial^2 C(\omega)}{\partial \omega_1 \partial \omega_2} & \cdots & \frac{\partial^2 C(\omega)}{\partial \omega_1 \partial \omega_p} \\
		\frac{\partial^2 C(\omega)}{\partial \omega_2 \partial \omega_1} & \frac{\partial^2 C(\omega)}{\partial^2 \omega_2} & \cdots & \frac{\partial^2 C(\omega)}{\partial \omega_2 \partial \omega_p} \\				
		\vdots & \vdots & \ddots & \vdots \\
		\frac{\partial^2 C(\omega)}{\partial \omega_p \partial \omega_1} & \frac{\partial^2 C(\omega)}{\partial \omega_p \partial \omega_2} & \cdots & \frac{\partial^2 C(\omega)}{\partial^2 \omega_p}			
	\end{bmatrix}_{\omega=\hat{\omega}} \in\mathbb{R}^{P\times P},\quad
	v = \begin{bmatrix}
		v_1 \\
		v_2 \\
		\vdots \\
		v_P
	\end{bmatrix}\in\mathbb{R}^P,
\end{equation}
where $C(\omega)$ is the scalar cost function \eqref{eq:cost}, $\omega\in\mathbb{R}^P$ denotes the model parameter vector, and where $\hat{\omega}$ is the point in parameter space where we would like to evaluate the Hessian. The implementation of \texttt{get\_Hv\_op()} in \texttt{pyhessian} is as follows
\begin{lstlisting}[basicstyle=\scriptsize,mathescape=true,frame=tb, caption={get\_Hv\_op() implementation}, label={lst:label9}]
y = flatten(tf.gradients(tf.math.multiply(flatten(tf.gradients(C, $\hat{\omega}$)), 
                                          tf.stop_gradient(v)), 
                         params))
\end{lstlisting}

The inner-most differentiation (e.g. \texttt{tf.gradients()}) will return the gradient of the scalar function $C(\omega)$ evaluated at $\omega=\hat{\omega}$, which we will denote by $\nabla_{\omega} C(\omega)|_{\omega=\hat{\omega}}\in\mathbb{R}^P$. Furthermore, this gradient is multiplied element-wise by the vector $v$, and we get
\begin{equation}
	\nabla_{\omega} C(\omega) \circ v|_{\omega=\hat{\omega}} = \begin{bmatrix}
		\frac{\partial C(\omega)}{\partial \omega_1} v_1 \\
		\frac{\partial C(\omega)}{\partial \omega_2} v_2 \\
		\vdots \\
		\frac{\partial C(\omega)}{\partial \omega_P}v_P
	\end{bmatrix}_{\omega=\hat{\omega}}.
\end{equation}
Therefore the first argument of the outer-most differentiation (e.g. \texttt{tf.gradients()}), will be a vector function rather than a scalar function as was not the case in the inner-most differentiation. Since differentiation of tensors in TensorFlow will evaluate to the \underline{sum} of the gradients of the individual elements (of the tensor which is differentiated), we get

\begin{align}
	\nabla_{\omega} \nabla_{\omega} C(\omega) \circ v|_{\omega=\hat{\omega}} &= \begin{bmatrix} \frac{\partial}{\partial \omega_1}\frac{\partial C(\omega)}{\partial \omega_1} v_1 + \frac{\partial}{\partial \omega_1}\frac{\partial C(\omega)}{\partial \omega_2} v_2 + \hdots + \frac{\partial}{\partial \omega_1}\frac{\partial C(\omega)}{\partial \omega_P} v_P \\
		\frac{\partial}{\partial \omega_2}\frac{\partial C(\omega)}{\partial \omega_1} v_1 + \frac{\partial}{\partial \omega_2}\frac{\partial C(\omega)}{\partial \omega_2} v_2 + \hdots + \frac{\partial}{\partial \omega_2}\frac{\partial C(\omega)}{\partial \omega_P} v_P \\ 
		\vdots \\
		\frac{\partial}{\partial \omega_P}\frac{\partial C(\omega)}{\partial \omega_1} v_1 + \frac{\partial}{\partial \omega_P}\frac{\partial C(\omega)}{\partial \omega_2} v_2 + \hdots + \frac{\partial}{\partial \omega_P}\frac{\partial C(\omega)}{\partial \omega_P} v_P \\	
	\end{bmatrix}_{\omega=\hat{\omega}}\\
	&= \begin{bmatrix} \frac{\partial^2 C(\omega)}{\partial^2 \omega_1} v_1 + \frac{\partial^2 C(\omega)}{\partial \omega_1 \partial \omega_2} v_2 + \hdots + \frac{\partial^2 C(\omega)}{\partial \omega_1 \partial \omega_P} v_P \\
		\frac{\partial^2 C(\omega)}{\partial \omega_2 \partial \omega_1} v_1 + \frac{\partial^2 C(\omega)}{\partial^2 \omega_2} v_2 + \hdots + \frac{\partial^2 C(\omega)}{\partial \omega_2 \partial \omega_P} v_P \\ \vdots \\
		\frac{\partial^2 C(\omega)}{\partial \omega_P \partial \omega_1} v_1 + \frac{\partial^2 C(\omega)}{\partial \omega_P \partial \omega_2} v_2 + \hdots + \frac{\partial^2 C(\omega)}{\partial^2 \omega_P} v_P \\
	\end{bmatrix}_{\omega=\hat{\omega}} \\
	&= \begin{bmatrix} 
		\frac{\partial^2 C(\omega)}{\partial^2 \omega_1} & \frac{\partial^2 C(\omega)}{\partial \omega_1 \partial \omega_2} & \cdots & \frac{\partial^2 C(\omega)}{\partial \omega_1 \partial \omega_p} \\
		\frac{\partial^2 C(\omega)}{\partial \omega_2 \partial \omega_1} & \frac{\partial^2 C(\omega)}{\partial^2 \omega_2} & \cdots & \frac{\partial^2 C(\omega)}{\partial \omega_2 \partial \omega_p} \\				
		\vdots & \vdots & \ddots & \vdots \\
		\frac{\partial^2 C(\omega)}{\partial \omega_p \partial \omega_1} & \frac{\partial^2 C(\omega)}{\partial \omega_p \partial \omega_2} & \cdots & \frac{\partial^2 C(\omega)}{\partial^2 \omega_p}			
	\end{bmatrix}
	\begin{bmatrix}
		v_{1} \\
		v_{2} \\
		\vdots \\
		v_{P}
	\end{bmatrix}_{\omega=\hat{\omega}}\\
	&= v^TH(\omega)|_{\omega=\hat{\omega}}\qquad\QED
\end{align}
\subsection{Derivation of the Full-rank Approximation}\label{sec:full-rankproof}
The full eigendecomposition of the Hessian matrix can be written
\begin{equation}
	H = Q_L \Lambda_L Q_L^T + Q_R \Lambda_R Q_R^T,
\end{equation}
where $Q_L \in \mathbb{R}^{P \times K}$ is the matrix whose $k$th column is the eigenvector $q_k$ of $H$, and $\Lambda_L \in \mathbb{R}^{K \times K}$ is the diagonal matrix whose elements are the corresponding eigenvalues, ${\Lambda_L}_{kk} = \lambda_k$. Further, $Q_R \in \mathbb{R}^{P \times (P-K)}$ is the matrix whose $k$th column is the eigenvector $q_{K+k}$ of $H$, and $\Lambda_R \in \mathbb{R}^{(P-K) \times (P-K)}$ is the diagonal matrix whose elements are the corresponding eigenvalues, ${\Lambda_R}_{kk} = \lambda_{K+k}$. We assume that the eigenvalues are algebraically sorted so that $\lambda_1 \ge \lambda_2 \ge \lambda_K \ge \hdots \ge \lambda_P$. Assuming that the eigenvalues $\lambda_{K+1} = \lambda_{K+2} = \hdots = \lambda_{P} = \widetilde{\lambda} > 0$, we get
\begin{align}
	\widetilde{\widetilde{H}} &= Q_L \Lambda_L Q_L^T + Q_R \widetilde{\lambda}I Q_R^T\\
	  &= Q_L \Lambda_L Q_L^T + \widetilde{\lambda}Q_R Q_R^T.
\end{align}
Since the columns of $Q_L$ and $Q_R$ forms an orthonormal basis, it follows that $I = Q_L Q_L^T + Q_R Q_R^T$, and thus
\begin{align}
	\widetilde{\widetilde{H}} &= Q_L \Lambda_L Q_L^T + \widetilde{\lambda}(I - Q_L Q_L^T).
\end{align}
Consequently, $\widetilde{\widetilde{H}}$ will be full-rank since all its eigenvalues are greater than zero.\QED
\begin{multicols}{2}

\end{multicols}
\end{document}